\documentclass[letterpaper]{article}
\usepackage{aaai2027}  
\usepackage{times}  
\usepackage{helvet}  
\usepackage{courier}  
\usepackage[hyphens]{url}  
\usepackage{graphicx} 
\usepackage{natbib}  
\usepackage{caption} 
\usepackage{booktabs}       
\usepackage{amsfonts}       
\usepackage{nicefrac}       
\usepackage{microtype}      
\usepackage{xcolor}         
\usepackage[table]{xcolor}
\usepackage{bm}
\usepackage{amsmath}
\usepackage{amssymb}
\usepackage{amsthm}         
\usepackage{makecell}
\usepackage[ruled,linesnumbered]{algorithm2e}
\usepackage[most]{tcolorbox}
\usepackage{adjustbox}
\usepackage{dashrule}

\providecommand{\uline}[1]{\underline{#1}}

\definecolor{bestcolor}{RGB}{210,228,245}
\definecolor{secondcolor}{RGB}{255,232,195}

\newcommand{\bestval}[1]{\cellcolor{bestcolor}$\bm{#1}$}
\newcommand{\secondval}[1]{\cellcolor{secondcolor}$\bm{#1}$}

\DontPrintSemicolon
\SetAlgoNoLine
\SetKwFor{For}{for}{do}{end for}

\definecolor{forceyellow}{HTML}{FFF3BF}
\definecolor{latentpurple}{HTML}{EDE4FF}
\definecolor{sectiongray}{HTML}{E8E8E8}

\title{Stochastic Control Policies for Robust Molecular Transition Path Sampling}

\author{
    Jingqian Liu\textsuperscript{\rm 1}, Yu-Hsiang Wang\textsuperscript{\rm 2}, Yanru Qu\textsuperscript{\rm 1}, Ge Liu\textsuperscript{\rm 1}
}
\affiliations{
    \textsuperscript{\rm 1}Siebel School of Computing and Data Science,
    University of Illinois Urbana--Champaign\\
    \textsuperscript{\rm 2}Department of Electrical and Computer Engineering,
    University of Illinois Urbana--Champaign\\
    Urbana, IL, 61801\\
    \{jl126, yw121, yanruqu2, geliu\}@illinois.edu
}
\nocopyright
\begin{document}

\maketitle

\begin{abstract}
Transition path sampling (TPS) aims to efficiently generate rare molecular transition trajectories between metastable states and is essential for understanding biomolecular mechanisms. Beyond traditional molecular dynamics (MD)-based sampling, machine learning
has become central to state-of-the-art TPS. One major class of methods learns control forces during explicit MD rollouts. By preserving the underlying molecular dynamics, these methods tend to produce more physically plausible trajectories than endpoint-conditioned generators that construct paths directly. However, rollout-based control methods have been reported to exhibit unstable and strongly seed-dependent performance. We recast rollout-based control as learning a path-space proposal distribution
and investigate stochasticity placement as a design choice for improving
exploration and optimization robustness.
We develop two stochastic policies: FS-TPS, which
directly parameterizes a state-dependent Gaussian distribution over the control policy output, and LaS-TPS, which samples a compact latent control variable and decodes it into structured, cross-atom-correlated force variation. We conduct extensive multi-seed experiments on three biomolecular systems of increasing size: alanine dipeptide, chignolin, and BBL, a fast-folding protein. Stochastic policies consistently improve transition success and path quality over deterministic-policy baselines while substantially reducing sensitivity to random initialization.

\end{abstract}

\section{Introduction}

Molecular transitions between metastable states underlie processes such as
protein folding, conformational switching, and chemical reactions. Transition
path sampling (TPS) aims to generate reactive trajectories connecting
metastable states, providing mechanistic insight into how biomolecular systems
evolve between long-lived conformations~\cite{dellago1998,bolhuis2002}. Efficiently sampling such trajectories is challenging because the relevant free-energy barriers are rarely crossed on accessible simulation timescales. Although carefully designed collective variables (CVs) can facilitate
this exploration, identifying informative CVs often requires substantial domain
knowledge. This limitation becomes more severe for protein folding, which involves high-dimensional and coordinated atomic
motions that may not admit an obvious low-dimensional representation.

Existing TPS approaches can be broadly divided into three families. Traditional
molecular dynamics (MD)-based methods~\cite{frenkel2002understanding} generate trajectories through molecular force fields,
sometimes aided by externally applied steering forces~\cite{smd} or biased sampling approaches~\cite{torrie1977umbrella,laio2002metadynamics}. Learning-guided MD-rollout
methods instead train a neural policy to predict an instantaneous bias from the
current molecular configuration. The bias is combined with the physical force,
and the system is advanced by the MD integrator, allowing the complete path to
emerge sequentially from controlled dynamics. PIPS~\cite{pips} introduced this
policy-learning formulation; TPS-DPS improved off-policy training with a replay
buffer and a log-variance path-measure objective
~\cite{tpsdps}. The third family,
including Doob's Lagrangian~\cite{doob} and FAS~\cite{fas}, directly constructs complete
endpoint-conditioned paths without MD rollouts. Among these families, learning-guided MD rollouts
remain competitive because they preserve explicit temporal propagation and tend
to generate more physically plausible paths than direct endpoint-conditioned
generators. Evidence reported in the Appendix shows that Doob's Lagrangian paths collapse toward a geometrical connection in the CV space for alanine dipeptide, consistent with its failure independently reported elsewhere~\citep{karczewski2026spacetime}. FAS produces less favorable low-energy paths on the larger benchmark, chignolin. However, rollout-based methods also have a clear
limitation: a poor initial replay buffer can repeatedly expose the policy to a
narrow region of path space, reinforce unsuccessful control strategies, and
produce strong sensitivity to network initialization.

Prior work has attempted to address this instability through trust-region
regularization~\cite{trlv}, but achieved only limited performance gains. In
this work, we explore a different perspective: introducing stochasticity into
policy learning. This connects to a classical result in stochastic
optimal control, where the noise level governs whether the optimal control
averages over multiple solutions or collapses onto a single one
\cite{kappen2005path,theodorou2010generalized}. Our motivation also comes from reinforcement learning, which shares
with MD-rollout methods a sequential control structure, state-dependent actions,
trajectory-based interaction, and replay-buffer-based off-policy optimization,
despite using a different learning objective. Previous studies have shown that action-space perturbations, parameter-space
noise, and entropy-regularized stochastic policies can promote exploration
\citep{lillicrap2015continuous,plappert2017parameter,fortunato2017noisy,haarnoja2018soft},
grounded in the broader connection between maximum-entropy control and
probabilistic inference~\citep{levine2018reinforcement}. Inspired by these approaches, we first
replace the deterministic bias-force predictor with a state-dependent Gaussian
policy that samples the applied control during each MD rollout, yielding
force-space stochastic TPS (FS-TPS). On alanine dipeptide, FS-TPS improves
RMSD, transition success, and path diversity by $39.5\%$, $52.8\%$, and
$22.2\%$, respectively, relative to the deterministic baseline. These gains
remain consistent on the larger chignolin and BBL systems. More importantly,
evaluations over seven to nine random seeds show substantially reduced
variability, demonstrating greater robustness to network initialization.

These results further motivated us to examine where stochasticity should be introduced
in higher-dimensional molecular systems. Independent fluctuations in Cartesian
force components may be insufficient to represent the coordinated atomic motions
involved in molecular transitions. We therefore hypothesize that stochasticity
should instead be introduced through a compact latent representation, allowing a
shared decoder to transform latent perturbations into correlated variations of
the full atom-wise bias force. Based on this idea, we propose latent-space
stochastic TPS (LaS-TPS), which samples a state-dependent latent variable within
a bottleneck architecture and deterministically decodes it into the applied
control. Under a first-order local approximation, we show that the induced force
covariance is generally non-diagonal and state-dependent, with rank
upper-bounded by the latent dimension. Our empirical characterization further
shows that LaS-TPS concentrates stochastic variation into low-effective-rank,
cross-atom-correlated force modes, supporting the intended structured
parameterization. Using the same set of random
initializations, LaS-TPS reduces RMSD and transition-state energy by $41.9\%$
and $25.4\%$, respectively, while improving transition success and mode coverage
by $58.9\%$ and $77.8\%$ on alanine dipeptide relative to the deterministic
baseline. It also retains improvements in transition success and energetic path
quality on chignolin and BBL, with the gains evaluated over seven to nine random
seeds.\\
\smallskip
Our main contributions are as follows:
\begin{itemize}
    \item We revisit learning-guided MD rollouts from a path-space proposal
    perspective and introduce stochastic control as a mechanism for improving
    exploration and reducing sensitivity to early replay-buffer trajectories.

    \item We propose FS-TPS, which replaces the
    deterministic bias predictor with a state-dependent Gaussian control policy
    and uses entropy regularization to maintain exploration during off-policy
    training. Across alanine dipeptide, chignolin, and BBL, FS-TPS improves
    transition success and reduces sensitivity to network initialization over
    seven to nine random seeds.

    \item We propose LaS-TPS, which introduces
    stochasticity through a compact bottleneck representation. Under local
    linearization, we show that the induced force covariance is generally
    correlated and state-dependent, with rank upper-bounded by the latent
    dimension, and empirically characterize these structured force variations.
    Across the three molecular systems, LaS-TPS achieves the strongest overall
    transition-success and energetic performance while maintaining robustness over random seeds.
\end{itemize}

\section{Related work}
\paragraph{Monte Carlo path sampling and learned proposals.}
Transition path sampling was originally formulated as Monte Carlo sampling in trajectory space, where shooting and shifting moves generate new reactive trajectories from existing ones~\citep{dellago1998,bolhuis2002}. The efficiency of these samplers depends critically on the quality of the shooting points, which motivated a line of work that replaces hand-designed moves with learned proposal distributions. Boltzmann generators~\citep{boltzmanngen} and their conditional variants~\citep{condbg} learn to sample equilibrium configurations directly, and have been used to propose shooting points or MCMC moves in path space~\citep{bgmcmc}. Related approaches guide path sampling with learned reaction coordinates~\citep{jung2023}, allocate simulation effort adaptively through reinforcement learning~\citep{rlwe}, or employ diffusion models as enhanced-sampling proposals for rare events and free-energy estimation~\citep{eds}. 
\paragraph{Rollout-based learned control.}
A second line of work replaces the Monte Carlo machinery with a learned biasing force that steers the dynamics toward the target basin during explicit molecular-dynamics rollouts. The reactive path ensemble was first cast as a variational control problem in this form, with a time-dependent control force reweighting the unbiased path measure~\citep{singh2023}. PIPS~\citep{pips} introduced a collective-variable-free policy-learning perspective, but its KL-based objective can suffer from mode collapse, capturing only a subset of the reactive path modes. TPS-DPS~\citep{tpsdps} formulates transition path generation as a controlled diffusion process and optimizes the policy with a log-variance objective~\citep{nusken2021}, alleviating mode collapse but remaining sensitive to random seeds and policy initialization. TR-LV~\citep{trlv} adds trust-region constraints to log-variance optimization, substantially improving stability while yielding only moderate gains in transition-path metrics. Throughout this line, paths are generated by sequential MD rather than imposed by endpoint construction, preserving a closer connection to physically time-ordered dynamics. All of these methods, however, learn a \emph{deterministic} map from configuration to control force: the policy commits to a single control action per configuration.
\paragraph{Endpoint-conditioned and path-space generative models.}
A complementary family constructs entire trajectories at once, enforcing the two metastable endpoints by construction rather than reaching the target basin through simulation. One group formulates TPS through Doob's $h$-transform, which characterizes stochastic processes conditioned on rare events: Doob's Lagrangian~\citep{doob} learns an endpoint-conditioned path distribution through a simulation-free variational objective, and subsequent work scales this formulation to longer transition paths with a sequence-to-sequence architecture and fixed-window attention~\citep{doobseq}. A second group generates the trajectory with a diffusion or flow model, by minimizing the Onsager--Machlup action~\citep{onsagermachlup}, learning normalizing flows over paths~\citep{pathflow}, or training directly on trajectory data~\citep{dynamicsdiffusion,diffusiontp,genflowmatching}. FAS~\citep{fas} extends adjoint-based sampling~\citep{pis,richter2024,havens2025adjoint} to function spaces, sampling endpoint-constrained path functions with a terminal cost combining potential energy and path smoothness.

\section{Methodology}

\subsection{Log-variance objective in learning-guided transition path sampling}

TPS-DPS~\cite{tpsdps} trains a biasing policy by minimizing a log-variance objective between the path distribution induced by biased molecular dynamics and the target transition-path distribution. Because this objective can be evaluated on trajectories stored in a replay buffer, it supports off-policy training and improves sample efficiency. A learnable control variate estimates the unknown normalization constant and reduces the variance of the gradient estimator. We refer readers to the original TPS-DPS paper~\cite{tpsdps} for the complete derivation and training procedure.

The deterministic TPS-DPS policy takes the current molecular configuration and target conformation as input and produces the bias applied during molecular-dynamics rollouts. Two output parameterizations are considered. The \emph{force-based} parameterization directly predicts the Cartesian bias force. In contrast, the \emph{scale-based} parameterization predicts atom-wise scaling coefficients, which are multiplied by the displacement toward the aligned target conformation to construct the bias force. The latter introduces a target-directed inductive bias and is more scalable for larger molecular systems. Following the original system-specific setup, we use direct force outputs for alanine dipeptide and scale outputs for the larger protein systems. All proposed stochastic policy variants use the same output parameterization as their corresponding deterministic TPS-DPS baseline to ensure a controlled comparison.

\subsection{Force-Space Distributional Control (FS-TPS)}

For FS-TPS, we relax the assumption that the policy is
deterministic by modeling the policy as a conditional
probability distribution over admissible bias forces. Given
the current configuration $x_t$ and target conformation
$x_{\mathcal B}$, the policy defines a diagonal Gaussian
distribution
\begin{equation}
\pi_\theta(\mathbf{u}_t\mid x_t, x_{\mathcal B})
=
\mathcal N
\left(
\boldsymbol{\mu}_\theta(x_t, x_{\mathcal B}),
\operatorname{diag}\!\left(\boldsymbol{\sigma}_\theta^2(x_t, x_{\mathcal B})\right)
\right),
\end{equation}

where $\boldsymbol{\mu}_\theta(x_t, x_{\mathcal B})$ and
$\boldsymbol{\sigma}_\theta(x_t, x_{\mathcal B})$ denote the
predicted mean and standard deviation of the bias force,
respectively. A bias force is sampled via the reparameterization
trick~\cite{kingma2014vae,rezende2014stochastic}.
\begin{equation}
\mathbf{u}_t
=
\boldsymbol{\mu}_\theta(x_t, x_{\mathcal B})
+
\boldsymbol{\sigma}_\theta(x_t, x_{\mathcal B})
\odot
\boldsymbol{\epsilon},
\qquad
\boldsymbol{\epsilon}
\sim
\mathcal N(\mathbf0,\mathbf I),
\end{equation}

\subsubsection{Entropy-Regularized Path-Measure Objective}
We
regularize the controller by maximizing the conditional differential
entropy of the bias-force policy. For the diagonal Gaussian policy,
the entropy at molecular state $x_t$ is
\begin{equation}
\mathcal{H}
\left(
\pi_\theta(\cdot \mid x_t, x_{\mathcal B})
\right)
=
\frac{1}{2}
\sum_{i=1}^{d_u}
\log
\left(
2\pi e\,
\sigma_{\theta,i}^{2}(x_t, x_{\mathcal B})
\right),
\end{equation}
where $d_u$ denotes the dimensionality of the bias force. We define the
trajectory-averaged policy entropy as
\begin{equation}
\overline{\mathcal{H}}(\pi_\theta)
=
\mathbb{E}_{\tau}
\left[
\frac{1}{T}
\sum_{t=0}^{T-1}
\mathcal{H}
\left(
\pi_\theta(\cdot \mid x_t, x_{\mathcal B})
\right)
\right].
\end{equation}

The final training objective combines path-measure matching ($\mathcal{L}_{\mathrm{LV}}$) with
entropy regularization,

\begin{equation}
\mathcal{L}_{\mathrm{FS}}
=
\mathcal{L}_{\mathrm{LV}}
-
\lambda_{\mathcal{H}}
\overline{\mathcal{H}}(\pi_\theta).
\end{equation}

This objective encourages the induced path measure to approach the
target transition-path measure while preventing premature collapse of
the state-dependent bias-force distribution. To account for variations in thermal fluctuations across simulation
conditions, we use a temperature-dependent coefficient with a tunable
hyperparameter $\lambda_{\mathcal{H}}^{\mathrm{ref}}$, where
$T_{\mathrm{ref}}$ is the final temperature of the annealing schedule
when one is used, and the training temperature otherwise (see Appendix
for details).
\begin{equation}
\lambda_{\mathcal{H}}(T)
=
\lambda_{\mathcal{H}}^{\mathrm{ref}}
\frac{T}{T_{\mathrm{ref}}},
\end{equation}

\subsection{Latent-Space Stochastic Control (LaS-TPS)}

LaS-TPS represents uncertainty through a compact stochastic
control variable. It adopts a variational
information bottleneck (VIB)~\cite{vib} architecture, in which an input is mapped to
a distribution over latent representations and a sample from this
distribution is used to perform the downstream task. Given the current configuration $x_t$ and target conformation
$x_{\mathcal B}$, the encoder predicts a diagonal Gaussian distribution
\begin{equation}
q_{\theta_{\mathrm e}}
\left(
\mathbf z_t \mid x_t,x_{\mathcal B}
\right)
=
\mathcal{N}\!\left(
\boldsymbol{\mu}_{\theta_{\mathrm e}}(x_t,x_{\mathcal B}),
\operatorname{diag}\!\left[
\boldsymbol{\sigma}_{\theta_{\mathrm e}}^2
(x_t,x_{\mathcal B})
\right]
\right).
\label{eq:las-latent}
\end{equation}
At each molecular-dynamics step, a latent control variable is sampled
using the reparameterization trick and decoded into the bias force:
\begin{equation}
\boldsymbol{\epsilon}_t
\sim\mathcal{N}(\mathbf 0,\mathbf I),
\qquad
\mathbf z_t
=
\boldsymbol{\mu}_{\theta_{\mathrm e}}
+
\boldsymbol{\sigma}_{\theta_{\mathrm e}}
\odot\boldsymbol{\epsilon}_t,
\qquad
\mathbf u_t
=
g_{\theta_{\mathrm d}}(\mathbf z_t).
\label{eq:las-sampling}
\end{equation}

The latent distribution is regularized toward a standard Gaussian
prior. The resulting LaS-TPS objective is
\begin{equation}
\small
\begin{aligned}
&\mathcal{L}_{\mathrm{LaS}}(\theta,\log Z)
=
\mathcal{L}_{\mathrm{LV}}(\theta,\log Z)
+
\\
&\frac{\beta_{\mathrm{KL}}}{K(L+1)}
\sum_{k=1}^{K}
\sum_{t=0}^{L}
D_{\mathrm{KL}}\!\left(
q_{\theta_{\mathrm e}}
\left(
\mathbf z_t \mid x_t^{(k)},x_{\mathcal B}
\right)
\,\middle\|\,
\mathcal{N}(\mathbf 0,\mathbf I)
\right).
\end{aligned}
\label{eq:las-objective}
\end{equation}
where coefficient $\beta_{\mathrm{KL}}$ controls the strength of the
latent information bottleneck (see Appendix for details). 

\paragraph{Replay-buffer sampling.}
For LaS-TPS, we use a mildly prioritized replay strategy in which approximately $5\%$ of the training samples are drawn from the $30\%$ of replay-buffer trajectories with the lowest RMSD, while the remaining samples follow the standard sampling procedure. An ablation study shows that it provides only a marginal improvement. Given its limited effect, we do not discuss it further in the main experimental section. Full implementation details and ablation results are provided in the Appendix.

\subsection{Proposition: Structured control covariance induced by latent stochasticity.}
In LaS-TPS, stochasticity is introduced in a low-dimensional latent
space and subsequently mapped by a shared decoder to the full
atom-wise control force. Consequently, a single latent perturbation may
simultaneously modify multiple atoms and Cartesian components, allowing
the policy to represent correlated control patterns such as coupled
atomic motion, compensating force directions, coordinated rotations,
and collective backbone displacement. The following proposition
characterizes the local covariance structure induced by this
parameterization.
\smallskip
\label{prop:latent-covariance}
Let the latent space
\begin{equation}
\begin{split}
z\mid x, x_{\mathcal B}
\sim
\mathcal N\!\left(
\mu_z(x, x_{\mathcal B}),
\Sigma_z(x, x_{\mathcal B})
\right),
\\
\qquad
\Sigma_z(x, x_{\mathcal B})
=
\operatorname{diag}\!\left(\sigma_z^2(x, x_{\mathcal B})\right),
\end{split}
\end{equation}
and the control force be given by
\begin{equation}
u=g_\theta(z),
\end{equation}
where $z\in\mathbb{R}^{d_z}$ and $u\in\mathbb{R}^{d_u}$. Define
\begin{equation}
J_z(x, x_{\mathcal B})
=
\left.
\frac{\partial g_\theta(z)}{\partial z}
\right|_{z=\mu_z(x,x_{\mathcal B})}.
\end{equation}
Under a first-order approximation around $z=\mu_z(x, x_{\mathcal B})$,
\begin{align}
\mathbb{E}[u\mid x, x_{\mathcal B}]
&\approx
g_\theta\!\left(\mu_z(x, x_{\mathcal B})\right),\\
\operatorname{Cov}(u\mid x, x_{\mathcal B})
&\approx
J_z(x, x_{\mathcal B})\Sigma_z(x, x_{\mathcal B})J_z(x, x_{\mathcal B})^\top.
\end{align}
Therefore, the induced force covariance is generally non-diagonal and
state-dependent, satisfies
\begin{equation}
\operatorname{rank}\!\left(
\operatorname{Cov}(u\mid x, x_{\mathcal B})
\right)
\leq d_z,
\end{equation}
and restricts the stochastic force variation to the column space of
$J_z(x, x_{\mathcal B})$. The proof is shown in the Appendix.

\paragraph{Empirical characterization of conditional force fluctuations.}
For each system, we construct a shared bank of fixed molecular configurations
spanning the transition from reactant to product
(\(S=200\) for alanine dipeptide and \(S=224\) for chignolin and BBL).
For each fixed state \(x\), the molecular coordinates are held constant and the
trained policy is queried \(K=1024\) times, producing samples from the
conditional bias-force distribution \(p_\theta(f\mid x, x_{\mathcal B})\).
Since the target conformation \(x_{\mathcal B}\) is held fixed throughout,
we omit it from the notation below.
The resulting variation reflects stochastic policy outputs at one exact
configuration, rather than variation across molecular states.
After mean-centering the sampled forces, we estimate
\[
\Sigma(x)
=
\frac{1}{K-1}
\sum_{k=1}^{K}
\left(f_k-\bar f\right)
\left(f_k-\bar f\right)^\top,
\]
and project out rigid-body translations and rotations to obtain the internal
covariance \(\Sigma_{\mathrm{int}}(x)\).
We summarize the structure of \(\Sigma_{\mathrm{int}}(x)\) for each fixed
state and average these summary statistics over the state bank to obtain one
value per trained checkpoint.
We additionally characterize latent-rank utilization and the variation of the
leading local force direction across states.
Full definitions and implementation details are provided in
the Appendix.

\section{Experiments}
\subsection{Experimental Settings}
We follow the TPS-DPS implementation frame and evaluation protocol~\cite{tpsdps} on three systems of
increasing size: alanine dipeptide (22 atoms), transitioning from C5 to C7ax,
and the fast-folding proteins chignolin (166 atoms) and BBL (711 atoms)~\citep{lindorfflarsen2011},
transitioning from unfolded to folded states. All paths are generated on the fly
through biased MD using the TPS-DPS simulation settings~\cite{openmm,amber,langevin}. Training, sampling, and evaluation are
performed independently for each random seed, and each inference run generates
64 paths. We report heavy-atom final-state RMSD after Kabsch
alignment~\cite{kabsch1976}, target hit percentage (THP), and transition-state
energy (ETS), the maximum potential energy among states in a transition path. For alanine dipeptide, we additionally report mode coverage (also referred to as path diversity) over successful paths to assess whether
they visit the two predefined saddle regions. Coverage is $0\%$, $50\%$, or $100\%$ when none,
one, or both regions are covered, respectively. Further details are provided
in the Appendix.

\subsection{FS-TPS improves training exploration and the learned mean control}

\begin{figure}[htbp]
    \centering
    \hspace*{-5mm}%
    \includegraphics[width=0.52\textwidth]{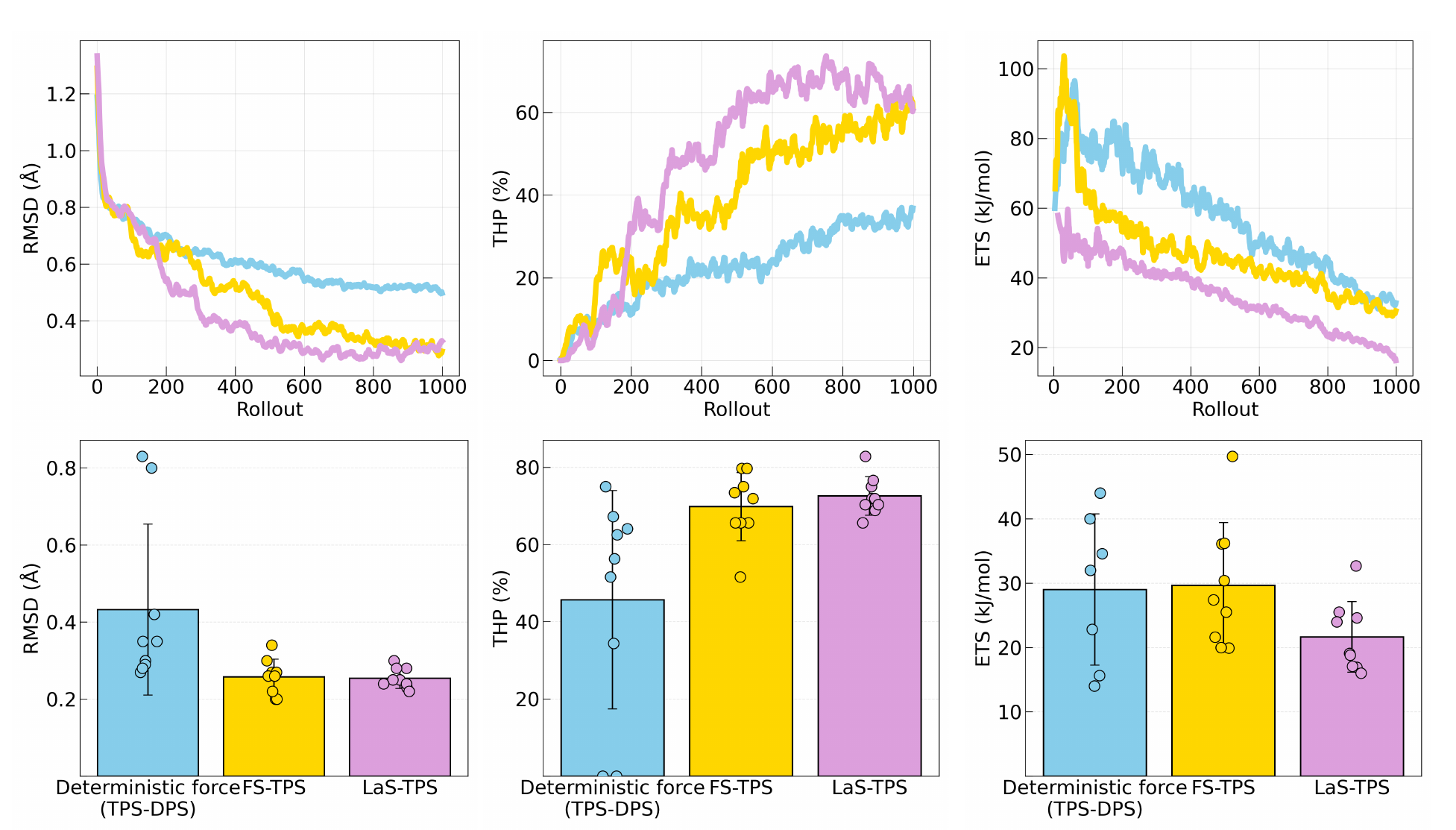}
    \caption{Training and inference performance of TPS-DPS, FS-TPS, and LaS-TPS on alanine dipeptide transition path sampling,
averaged over nine random seeds. \textbf{Top:} RMSD, THP, and ETS as a function
of training rollout, showing the evolution of each metric during training.
\textbf{Bottom:} Final RMSD, THP, and ETS at inference, with individual seed
results overlaid as points and error bars denoting standard deviation across seeds.}
    \label{fig:aldp-rollout}
\end{figure}

\begin{figure}[htbp]
    \hspace*{2mm}%
    \makebox[\columnwidth][c]{%
        \includegraphics[width=0.58\textwidth]
        {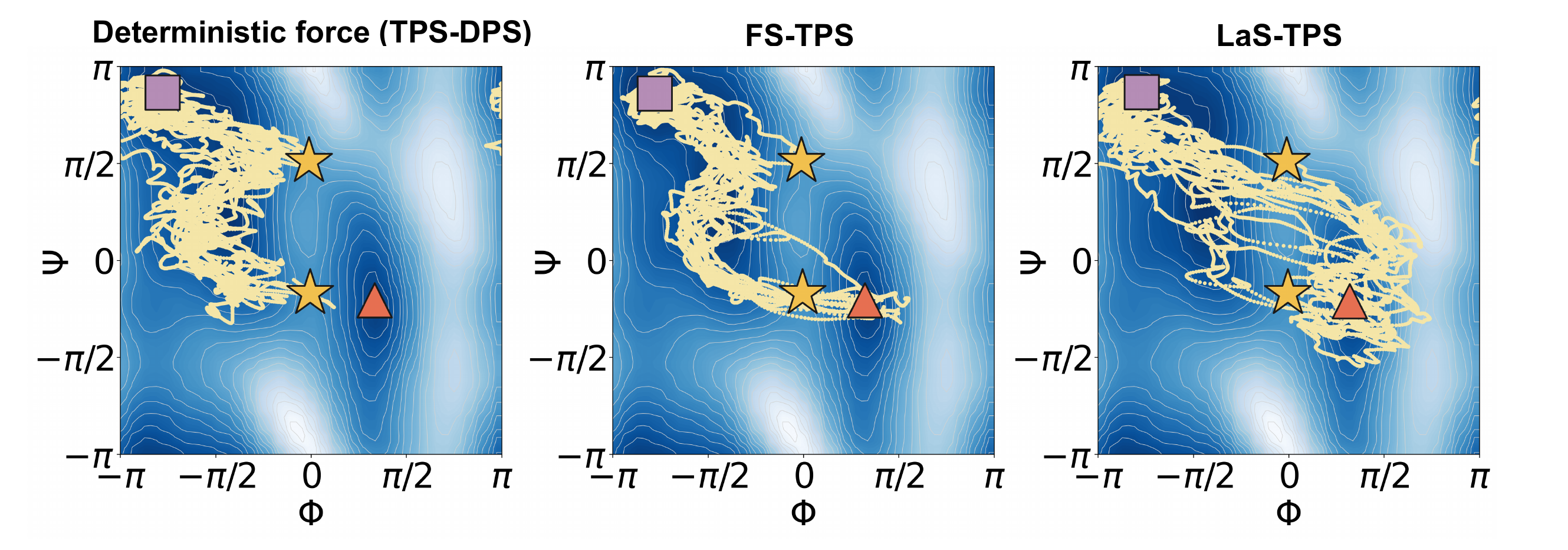}%
    }
    \caption{Comparison of sampled transition paths on the alanine
    dipeptide free-energy surface for TPS-DPS,
    FS-TPS, and LaS-TPS for a policy trained starting from the same random seed (72831). The stars show the two saddle regions.}
    \label{fig:aldp-phipsi}
\end{figure}

Figure~\ref{fig:aldp-rollout} shows that FS-TPS separates from the
deterministic baseline early in training, most clearly through a faster increase
in THP. Across nine seeds on alanine dipeptide, FS-TPS improves THP from
$45.67\pm28.26\%$ to $69.79\pm8.84\%$ and reduces RMSD from
$0.43\pm0.22$~\AA{} to $0.26\pm0.05$~\AA{}. The seed-level results in Figure~\ref{fig:aldp-phipsi} show how a run that remained
nearly unsuccessful under TPS-DPS is recovered by FS-TPS. Consistent improvements in THP are also observed on
chignolin and BBL, which is summarized in Table~\ref{tab:md-learning-comparison}.

We found entropy regularization is essential
to this behavior. Removing it largely degrades the performance. By directly discouraging premature variance collapse,
the entropy term maintains stochastic exploration while the replay buffer is
being constructed, as shown in Table~\ref{tab:ablation-aldp}.

Table~\ref{tab:ablation-aldp} also shows stochastic sampling is not required at inference for FS-TPS. Scaling
the learned standard deviation from $0$ to $1.5$ produces little change in RMSD
or THP. Only an extreme scale of $10$ substantially degrades ETS. These results suggest that
force-space stochasticity acts primarily during training by enriching early
rollouts and constructing a more successful replay buffer, thereby improving
the learned mean control rather than relying on inference-time noise.

We further examine whether simply adding temperature-scaled Gaussian noise to a deterministic policy can reproduce the THP achieved by FS-TPS. As reported in the Appendix, it fails to do so. These results indicate that the observed improvement cannot be attributed to additional randomness alone; rather, it arises from the learned, state-dependent force distribution sustained by entropy regularization.

\begin{table}[t]
\centering
\caption{
Ablation study of the proposed stochastic biasing policies on alanine
dipeptide. Unless otherwise indicated, results are reported as mean
$\pm$ standard deviation across the same nine random seeds.
ETS is computed over runs with valid transition paths.
Arrows indicate the preferred direction of each metric.
}
\label{tab:ablation-aldp}

\begingroup
\footnotesize
\setlength{\tabcolsep}{3pt}
\renewcommand{\arraystretch}{1.25}

\resizebox{1.0\linewidth}{!}{%
\begin{tabular}{@{}lccc@{}}
\toprule
\textbf{Method}
& \textbf{RMSD (\AA{} $\downarrow$)}
& \textbf{THP (\% $\uparrow$)}
& \textbf{ETS (kJ/mol $\downarrow$)} \\
\midrule

\rowcolor{gray!15}
\multicolumn{4}{@{}l}{\textit{Deterministic force control}} \\

\hspace{0.5em}TPS-DPS
& $0.43 \pm 0.22$
& $45.67 \pm 28.26$
& $29.00 \pm 11.75$ \\

\midrule
\rowcolor{gray!15}
\multicolumn{4}{@{}l}{\textit{Force-space stochastic control}} \\

\hspace{0.5em}FS-TPS ($\sigma$-scale $=1.0$)
& $0.26 \pm 0.05$
& $69.79 \pm 8.84$
& $29.64 \pm 9.76$ \\

\hspace{1.4em}$\sigma$-scale $=0$ (mean force)
& $0.25 \pm 0.04$
& $68.16 \pm 12.16$
& $31.64 \pm 10.36$ \\

\hspace{1.4em}$\sigma$-scale $=0.5$
& $0.26 \pm 0.06$
& $67.97 \pm 12.19$
& $31.35 \pm 8.01$ \\

\hspace{1.4em}$\sigma$-scale $=1.5$
& $0.25 \pm 0.05$
& $70.90 \pm 9.26$
& $32.21 \pm 9.64$ \\

\hspace{1.4em}$\sigma$-scale $=10$
& $0.26 \pm 0.05$
& $67.39 \pm 11.19$
& $49.93 \pm 7.44$ \\

\hspace{1.4em}w/o entropy regularization
& $0.54 \pm 0.22$
& $29.77 \pm 26.33$
& $43.68 \pm 38.48$ \\

\midrule
\rowcolor{gray!15}
\multicolumn{4}{@{}l}{\textit{Latent-space stochastic control}} \\

\hspace{0.5em}LaS-TPS ($\sigma$-scale $=1.0$)
& $0.25 \pm 0.03$
& $72.58 \pm 5.01$
& $21.63 \pm 5.49$ \\

\hspace{1.4em}$\sigma$-scale $=0$ (mean latent)
& $1.01 \pm 0.21$
& $0.98 \pm 1.16$
& $23.63 \pm 7.53$ \\

\hspace{1.4em}$\sigma$-scale $=0.5$
& $0.73 \pm 0.16$
& $8.59 \pm 5.48$
& $16.57 \pm 5.39$ \\

\hspace{1.4em}$\sigma$-scale $=1.5$
& $0.16 \pm 0.02$
& $88.87 \pm 4.90$
& $38.26 \pm 7.39$ \\

\hspace{1.4em}w/o KL regularization
& $0.25 \pm 0.07$
& $73.44 \pm 16.62$
& $29.40 \pm 13.21$ \\

\hspace{1.4em}Deterministic bottleneck
& $0.51 \pm 0.20$
& $37.13 \pm 29.74$
& $29.44 \pm 11.78$ \\

\bottomrule
\end{tabular}%
}
\endgroup
\end{table}

\subsection{From FS-TPS to LaS-TPS: Characterizing Sampled-Force Fluctuations}

\begin{table*}[t]
\centering
\caption{
The internal force covariance $\Sigma_{\mathrm{int}}(x)$ is characterized by
its effective rank, top-mode explained variance (EV@1), excess cross-atom
correlation relative to an atom-permuted null, and RMS stochastic amplitude
relative to the mean force.
For LaS-TPS, $r_{\mathrm{eff}}(z)/d_z$ measures latent-space rank utilization,
while global $r_{\mathrm{eff}}$ measures whether the dominant force direction
is shared across states.
Values are mean $\pm$ standard deviation over independently trained seeds.
}
\label{tab:stochastic-structure}

\begingroup
\setlength{\tabcolsep}{3pt}
\resizebox{0.7\textwidth}{!}{
\begin{tabular}{lcccccc}
\toprule
\textbf{Method}
& \textbf{Internal $r_{\mathrm{eff}}$}
& \textbf{Internal EV@1}
& \textbf{Excess $R_{\mathrm{cross}}$}
& \textbf{RMS noise/mean}
& \textbf{$r_{\mathrm{eff}}(z)/d_z$}
& \textbf{Global $r_{\mathrm{eff}}$} \\
\midrule

\multicolumn{7}{l}{\textbf{Alanine dipeptide}} \\

FS-TPS
& $52.56 \pm 3.00$
& $2.82\% \pm 0.16\%$
& $0.00 \pm 0.00$
& $0.20 \pm 0.05$
& /
& / \\

LaS-TPS
& $1.20 \pm 0.04$
& $95.86\% \pm 1.02\%$
& $0.58 \pm 0.02$
& $0.40 \pm 0.20$
& $0.98 \pm 0.00$ ($d_z=10$)
& $1.07 \pm 0.02$ \\

\midrule
\multicolumn{7}{l}{\textbf{Chignolin}} \\

FS-TPS
& $30.23 \pm 5.12$
& $12.30\% \pm 2.19\%$
& $0.00 \pm 0.00$
& $0.05 \pm 0.02$
& /
& / \\

LaS-TPS
& $1.00 \pm 0.00$
& $100.00\% \pm 0.00\%$
& $0.75 \pm 0.01$
& $1.34 \pm 0.06$
& $0.99 \pm 0.00$ ($d_z=24$)
& $14.18 \pm 1.33$ \\

\midrule
\multicolumn{7}{l}{\textbf{BBL}} \\

FS-TPS
& $139.09 \pm 6.57$
& $4.01\% \pm 0.34\%$
& $-0.0001 \pm 0.0002$
& $0.110 \pm 0.025$
& /
& / \\

LaS-TPS
& $1.204 \pm 0.094$
& $95.66\% \pm 2.01\%$
& $0.785 \pm 0.031$
& $1.133 \pm 0.108$
& $0.939 \pm 0.0002$ ($d_z=128$)
& $7.96 \pm 0.65$ \\

\bottomrule
\end{tabular}
}
\endgroup
\end{table*}

As established in the methodology section, a latent bottleneck allows a shared
decoder to transform compact stochastic perturbations into non-diagonal,
state-dependent force covariance with rank bounded by the latent dimension.
We use scalable MLPs with latent dimensions of 10, 24, and 128 for alanine
dipeptide, chignolin, and BBL, respectively.

To test the predicted structure, we characterize the conditional force
distributions over the shared transition-spanning state banks described in the
methodology section.
Local statistics are computed independently at each fixed configuration from
\(K=1024\) policy samples and then summarized over states and trained seeds.
Because the entropy-based effective rank is continuous, it may take non-integer
values.
As shown in Table~\ref{tab:stochastic-structure}, FS-TPS distributes its stochastic variation over many force directions, with internal effective ranks of \(52.56\), \(30.23\), and \(139.09\) for alanine dipeptide, chignolin, and BBL, respectively.
In contrast, LaS-TPS produces nearly one-dimensional local fluctuations, with
effective ranks close to one and a leading mode that explains almost all
internal-force variance.
These values are far below the corresponding latent-dimensional upper bounds
and do not arise from negligible stochastic amplitudes.

The near-unit output rank is not caused by collapse of the latent distribution.
We compute the effective rank of the latent covariance and normalize it by the
nominal latent dimension.
The resulting utilization ratios are close to one for all three systems,
indicating that most latent dimensions remain active.
Thus, the encoder does not reduce the latent distribution to a single active
coordinate; instead, the decoder maps broadly utilized latent variation into an
almost one-dimensional local force subspace. We further distinguish local dimensionality from variation of the dominant
direction across molecular states.
The global effective rank is computed by aggregating the sign-invariant outer
products of the leading local covariance directions across the state bank.
Alanine dipeptide has a global effective rank close to one, indicating that a
similar dominant direction is reused across configurations.
In contrast, chignolin and BBL have substantially larger global effective ranks.
Their fluctuations are therefore locally low-rank, while the dominant direction
changes across molecular configurations rather than collapsing globally to a
single shared force mode.

\begin{figure}[htbp]
    \hspace*{-5mm}%
    \centering
    \includegraphics[width=0.55\textwidth]{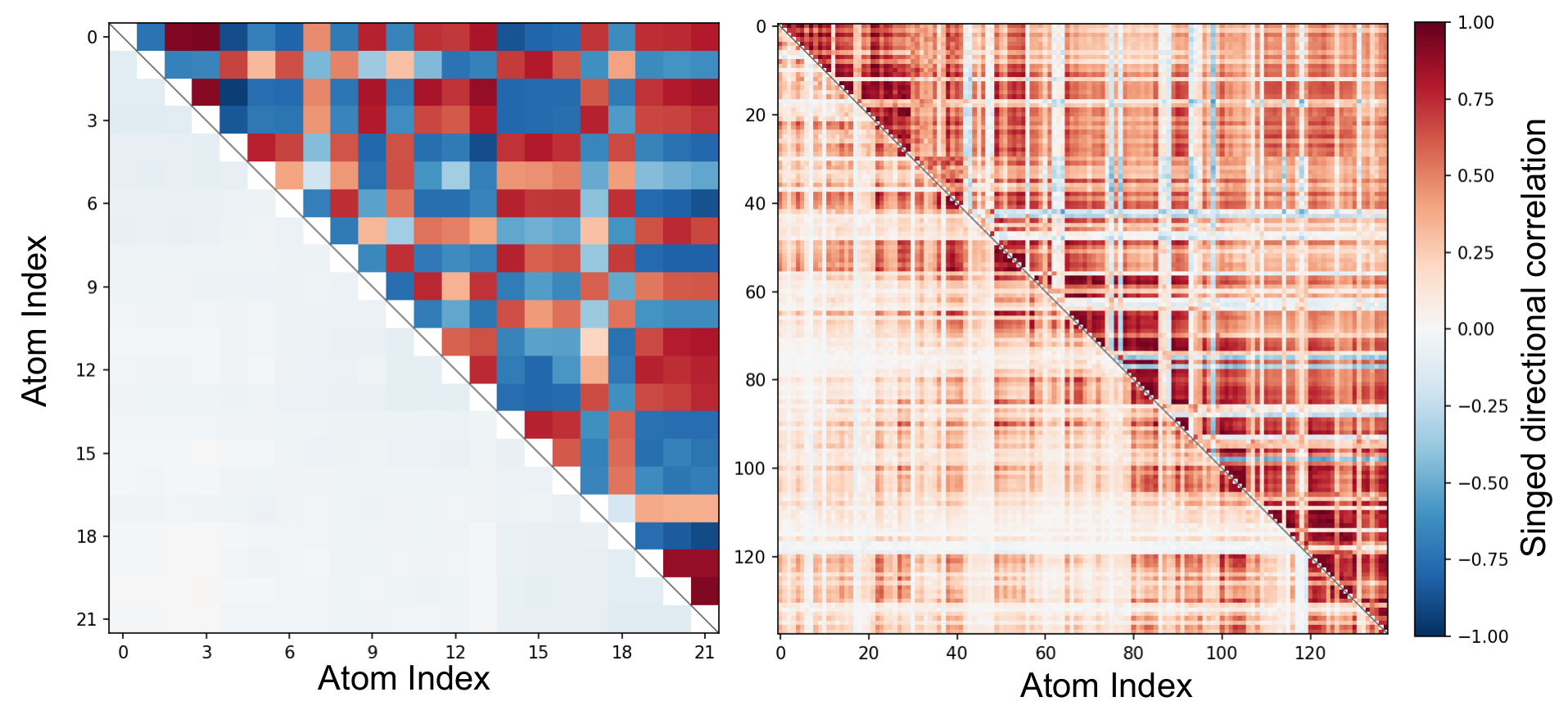}
    \caption{
Signed atom-pair directional correlations of the stochastic control for alanine dipeptide (left) and chignolin (right). The left lower and right upper triangles show FS-TPS and LaS-TPS, respectively. Positive and negative values indicate aligned and opposing force fluctuations. Results are averaged over seven training seeds.
}
\label{fig:noise_directional_correlation}
\end{figure}

In Table~\ref{tab:stochastic-structure}, we next quantify whether these fluctuations are coordinated across atoms.
For each atom pair, we measure the normalized magnitude of the corresponding
off-diagonal covariance block and subtract the same statistic under an
atom-permuted null, yielding the excess cross-atom correlation
\(R_{\mathrm{cross}}\). FS-TPS remains indistinguishable from the null across all three systems, whereas
LaS-TPS produces substantial excess correlation.
We also report the RMS stochastic amplitude relative to the mean force, which is
consistently larger for LaS-TPS than for FS-TPS.
Thus, LaS-TPS concentrates substantial force variation into a small number of
dominant directions rather than producing weak fluctuations with an apparently
low-dimensional covariance. Figure~\ref{fig:noise_directional_correlation} visualizes the signed directional correlations
for alanine dipeptide and chignolin, revealing strong cross-atom coordination
under LaS-TPS.
For chignolin, these correlations occur both among nearby atoms and across
several long-range atom pairs, showing that a single latent perturbation can
coordinate spatially separated molecular regions.

Overall, latent-space stochasticity is not merely a lower-dimensional source of
independent noise.
Instead, the decoder reorganizes latent variation into low-effective-rank,
cross-atom-correlated force fluctuations whose dominant direction remains
configuration dependent in the larger molecular systems.

\begin{table*}[t]
\centering
\caption{
Comparison of traditional MD and learning-guided transition path sampling
methods on alanine dipeptide, chignolin, and BBL.
Blue shading indicates the best reported result for each metric, whereas orange
shading indicates the second-best result among the methods included in this
table.
\textsuperscript{\dag} denotes results reported by Blessing et al.
Results for alanine dipeptide are averaged over the same nine random seeds.
Results for chignolin and BBL are averaged over the same seven random seeds.
\textbf{For BBL, the TPS-DPS statistics are computed over five successful runs; two
additional runs terminated during early rollouts because of numerical
instability.} $-$ indicates that implementation is not accessible while $/$ indicates that certain values are not applicable.
}
\label{tab:md-learning-comparison}

\begingroup
\fontsize{8.5}{9.4}\selectfont
\setlength{\tabcolsep}{1.0pt}
\renewcommand{\arraystretch}{1.04}

\begin{adjustbox}{max width=\textwidth,center}
\begin{tabular}{@{}lcccccccccc@{}}
\toprule
&
\multicolumn{4}{c}{Alanine Dipeptide (22 atoms)}
&
\multicolumn{3}{c}{Chignolin (166 atoms)}
&
\multicolumn{3}{c}{BBL (711 atoms)}
\\
\cmidrule(lr){2-5}
\cmidrule(lr){6-8}
\cmidrule(lr){9-11}

Method
& \makecell{RMSD\\(\AA{} $\downarrow$)}
& \makecell{THP\\(\% $\uparrow$)}
& \makecell{ETS\\(kJ/mol $\downarrow$)}
& \makecell{Mode cov.\\(\% $\uparrow$)}
& \makecell{RMSD\\(\AA{} $\downarrow$)}
& \makecell{THP\\(\% $\uparrow$)}
& \makecell{ETS\\(kJ/mol $\downarrow$)}
& \makecell{RMSD\\(\AA{} $\downarrow$)}
& \makecell{THP\\(\% $\uparrow$)}
& \makecell{ETS\\(kJ/mol $\downarrow$)}
\\
\midrule

\multicolumn{11}{@{}l}{\textit{Traditional MD}} \\

UMD\textsuperscript{\dag}
& $1.19 \pm 0.03$
& $6.25$
& $812.47 \pm 148.80$
& $0$
& $7.23 \pm 0.93$
& $1.56$
& $388.17$
& $18.48 \pm 0.63$
& $0.00$
& /
\\

SMD\textsuperscript{\dag}
& $0.56 \pm 0.27$
& $54.69$
& $78.40 \pm 12.76$
& \bestval{100}
& \bestval{0.85 \pm 0.24}
& $34.38$
& $179.52 \pm 138.87$
& $2.97 \pm 0.33$
& $7.81$
& $-1738.57 \pm 386.81$
\\

\midrule
\multicolumn{11}{@{}l}{\textit{Learning-guided transition path sampling}} \\

PIPS\textsuperscript{\dag}
& $0.66 \pm 0.15$
& $43.75$
& \secondval{28.17 \pm 10.86}
& $50$
& $4.66 \pm 0.17$
& $0.00$
& /
& $17.92 \pm 0.29$
& $0.00$
& /
\\

TPS-DPS
& $0.43 \pm 0.22$
& $45.67 \pm 28.26$
& $29.00 \pm 11.75$
& $50.00 \pm 43.30$
& $1.22 \pm 0.22$
& $27.47 \pm 16.88$
& $-530.85 \pm 171.38$
& $2.29 \pm 0.25$
& $28.76 \pm 13.19$
& $-2002.00 \pm 1892.27$
\\

TR-LV\textsuperscript{\dag}
& $0.29 \pm 0.03$
& $61.25 \pm 4.05$
& $49.11 \pm 5.84$
& --
& \secondval{0.90 \pm 0.01}
& \bestval{43.95 \pm 5.64}
& $-303.98 \pm 28.65$
& --
& --
& --
\\

FS-TPS (Ours)
& \secondval{0.26 \pm 0.05}
& \secondval{69.79 \pm 8.84}
& $29.64 \pm 9.76$
& $61.11 \pm 22.05$
& $1.16 \pm 0.15$
& $31.25 \pm 8.08$
& \secondval{-541.29 \pm 185.49}
& \bestval{1.81 \pm 0.20}
& \secondval{35.30 \pm 12.80}
& \secondval{-2282.71 \pm 718.30}
\\

LaS-TPS (Ours)
& \bestval{0.25 \pm 0.03}
& \bestval{72.58 \pm 5.01}
& \bestval{21.63 \pm 5.49}
& \secondval{88.89 \pm 22.05}
& $1.03 \pm 0.05$
& \secondval{38.84 \pm 10.63}
& \bestval{-629.00 \pm 76.88}
& \secondval{2.33 \pm 0.20}
& \bestval{36.39 \pm 14.21}
& \bestval{-2818.29 \pm 518.64}
\\

\bottomrule
\end{tabular}
\end{adjustbox}

\endgroup
\end{table*}

\begin{table}[t]
\centering
\caption{
Ablation study of the KL coefficient $\beta$ for LaS-TPS.
Results are reported as mean $\pm$ standard deviation across random seeds.
Arrows indicate the preferred direction of each metric.
Blue cells indicate the best result within each molecular system.
}
\label{tab:kl-beta-ablation}
 
\scriptsize
\setlength{\tabcolsep}{4pt}
\renewcommand{\arraystretch}{0.95}
 
\begin{tabular}{lccc}
\toprule
\textbf{$\beta$}
& \textbf{RMSD (\AA) $\downarrow$}
& \textbf{THP (\%) $\uparrow$}
& \textbf{ETS (kJ/mol) $\downarrow$} \\
\midrule
 
\rowcolor{sectiongray}
\multicolumn{4}{l}{\textbf{\textit{Alanine dipeptide}}} \\
 
$10^{-4}$
& $0.25 \pm 0.03$
& $72.58 \pm 5.01$
& \cellcolor{bestcolor}\textbf{$21.6 \pm 5.5$} \\
 
$10^{-3}$
& \cellcolor{bestcolor}\textbf{$0.23 \pm 0.05$}
& $73.69 \pm 6.99$
& $22.4 \pm 3.5$ \\
 
$10^{-2}$
& $0.27 \pm 0.07$
& \cellcolor{bestcolor}\textbf{$75.22 \pm 13.17$}
& $21.9 \pm 7.2$ \\
 
\midrule
 
\rowcolor{sectiongray}
\multicolumn{4}{l}{\textbf{\textit{Chignolin}}} \\
 
$10^{-4}$
& $1.21 \pm 0.18$
& $20.31 \pm 14.26$
& \cellcolor{bestcolor}\textbf{$-699.6 \pm 123.2$} \\
 
$10^{-3}$
& $1.03 \pm 0.05$
& $38.84 \pm 10.63$
& $-629.0 \pm 76.9$ \\
 
$10^{-2}$
& \cellcolor{bestcolor}\textbf{$0.99 \pm 0.12$}
& \cellcolor{bestcolor}\textbf{$46.43 \pm 12.68$}
& $-585.6 \pm 119.9$ \\
 
\midrule
 
\rowcolor{sectiongray}
\multicolumn{4}{l}{\textbf{\textit{BBL}}} \\
 
$10^{-7}$
& \cellcolor{bestcolor}\textbf{$2.14 \pm 0.29$}
& $25.89 \pm 11.71$
& \cellcolor{bestcolor}\textbf{$-3069.1 \pm 333.2$} \\
 
$10^{-6}$
& $2.33 \pm 0.20$
& \cellcolor{bestcolor}\textbf{$36.39 \pm 14.21$}
& $-2818.3 \pm 518.6$ \\
 
\bottomrule
\end{tabular}
\end{table}

\subsection{LaS-TPS Achieves the Strongest Overall Performance among MD-Rollout Methods}

Across the three molecular systems, LaS-TPS achieves the strongest overall
performance among the MD-rollout-based methods while remaining robust across
random seeds, as shown in Table~\ref{tab:md-learning-comparison}.
Its advantage is particularly pronounced in energetic path quality: LaS-TPS
obtains the lowest ETS among the rollout-based methods on alanine dipeptide,
chignolin, and BBL, while also maintaining competitive RMSD and transition
success.
On BBL, both FS-TPS and LaS-TPS alleviate the numerical instability observed in training of TPS-DPS. In the deterministic baseline, occasional excessively large predicted bias forces cause the MD simulation to diverge during early rollouts, whereas no such failures are observed for either stochastic policy. This improved stability is consistent with the stochastic parameterizations and their associated regularization producing less brittle control outputs and reducing the likelihood of extreme bias-force predictions.

These gains are not attributable to greater model capacity.
Across the three systems, LaS-TPS uses substantially fewer trainable parameters
than the corresponding baseline MLPs, amounting to only \(20.8\%\),
\(13.4\%\), and \(71.5\%\) of the baseline parameter counts for alanine
dipeptide, chignolin, and BBL, respectively.
Complete architecture specifications are provided in the Appendix. They also cannot be explained by the bottleneck architecture alone.
On alanine dipeptide, replacing the stochastic latent variable with a
deterministic bottleneck reduces THP below the TPS-DPS baseline and substantially
increases RMSD, indicating that simply inserting a low-dimensional hidden layer
does not recover the benefit of LaS-TPS.
Similarly, directly decoding the latent mean at inference yields poor endpoint
accuracy and almost no successful transitions.

KL regularization is also important for controlling the learned latent
distribution.
Results in Table~\ref{tab:ablation-aldp} show removing the KL term preserves endpoint performance on alanine
dipeptide but increases variability and degrades energetic path quality.
Moreover, varying the KL coefficient reveals a consistent trade-off between
transition success and ETS on the larger systems, as shown in Table~\ref{tab:kl-beta-ablation},
consistent with the
rate--distortion trade-off characterized by the information bottleneck
framework~\cite{tishby1999information,vib,betavae}.
A stronger KL penalty compresses the
configuration-dependent latent representation toward the shared prior, providing
greater regularization and often improving target-reaching robustness.
A weaker penalty retains more state-specific information, which can enable
lower-energy control strategies but may reduce transition success.
The KL coefficient controls not only latent regularization but also the
balance between reliably reaching the target basin and generating energetically
favorable transition paths. 

Because the conditional force covariance consistently exhibits an effective rank
close to one, we further tested whether a one-dimensional latent space would be
sufficient.
For alanine dipeptide, we reduced the latent dimension from \(10\) to \(1\) and
evaluated four random seeds.
One of them showed clear degradation, with a THP of \(21.9\%\) and an ETS of
\(60.3\,\mathrm{kJ/mol}\); full results are provided in the Appendix.
These results suggest that, although a scalar latent variable can be sufficient
when training succeeds, the higher-dimensional latent representation improves
optimization robustness and reduces sensitivity to initialization.

\section{Conclusion and Limitations}

We introduced two stochastic-control formulations for MD-rollout-based
TPS.
FS-TPS directly models a state-dependent distribution in force space, whereas
LaS-TPS injects stochasticity through a compact latent representation that is
decoded into the full atom-wise bias force.
Across three molecular systems, both approaches improve robustness
to random initialization and transition-path quality relative to deterministic
control, with LaS-TPS providing the strongest overall balance of transition
success and energetic quality.
Our analysis further shows that the location of stochasticity substantially
changes the resulting control distribution.
While FS-TPS produces broadly distributed, approximately factorized force
fluctuations, LaS-TPS organizes latent variation into low-effective-rank,
cross-atom-correlated force patterns whose dominant directions remain
configuration dependent in the larger systems.
These results suggest that high-dimensional molecular control may benefit not
only from introducing stochasticity, but also from learning how stochastic
variation is structured across atoms and force components. 

However, we found there is always a trade-off between THP and ETS.
Within LaS-TPS, varying the KL coefficient changes the balance between THP and
ETS.
A related trade-off appears more broadly across different TPS formulations.
MD-rollout-based methods must discover the target basin through sequential
simulation and therefore do not guarantee endpoint satisfaction, whereas
endpoint-conditioned generators achieve high transition success by construction
but may produce paths with less favorable energetic or dynamical properties.
Developing objectives that jointly balance transition probability, energetic
quality, and path diversity therefore remains an open challenge.
We hope that LaS-TPS will motivate further work on compact stochastic
representations that transform high-dimensional molecular uncertainty into
coordinated and scalable control strategies.

\bibliography{reference}

\clearpage
\appendix



\providecommand{\uline}[1]{\underline{#1}}

\definecolor{bestcolor}{RGB}{210,228,245}
\definecolor{secondcolor}{RGB}{255,232,195}

\DontPrintSemicolon
\SetAlgoNoLine
\SetKwFor{For}{for}{do}{end for}

\definecolor{forceyellow}{HTML}{FFF3BF}
\definecolor{latentpurple}{HTML}{EDE4FF}
\definecolor{sectiongray}{HTML}{E8E8E8}


\setcounter{secnumdepth}{0} 


\title{Appendix for\\
Stochastic Control Policies for Robust Molecular Transition Path Sampling}

\maketitle

\appendix
\section{Appendix}
\subsection{Proof of the Structured-Covariance Proposition}
\renewcommand{\qedsymbol}{}
Let
\[
u=g_\theta(z),
\qquad
z\mid x,x_{\mathcal B}
\sim
\mathcal N\!\left(
\mu_z(x,x_{\mathcal B}),
\Sigma_z(x,x_{\mathcal B})
\right).
\]
A first-order Taylor expansion of $g_\theta$ around the state-dependent
latent mean $\mu_z(x,x_{\mathcal B})$ gives
\begin{equation}
u
\approx
g_\theta\!\left(\mu_z(x,x_{\mathcal B})\right)
+
J_z(x,x_{\mathcal B})\left(z-\mu_z(x,x_{\mathcal B})\right),
\end{equation}
where
\begin{equation}
J_z(x,x_{\mathcal B})
=
\left.
\frac{\partial g_\theta(z)}{\partial z}
\right|_{z=\mu_z(x,x_{\mathcal B})}.
\end{equation}
Since
\begin{equation}
\mathbb{E}\!\left[z-\mu_z(x,x_{\mathcal B})\mid x,x_{\mathcal B}\right]=0,
\qquad
\operatorname{Cov}(z\mid x,x_{\mathcal B})=\Sigma_z(x,x_{\mathcal B}),
\end{equation}
the conditional mean and covariance are approximately
\begin{equation}
\mathbb{E}[u\mid x,x_{\mathcal B}]
\approx
g_\theta\!\left(\mu_z(x,x_{\mathcal B})\right),
\end{equation}
and
\begin{equation}
\operatorname{Cov}(u\mid x,x_{\mathcal B})
\approx
J_z(x,x_{\mathcal B})\Sigma_z(x,x_{\mathcal B})J_z(x,x_{\mathcal B})^\top.
\end{equation}
Moreover,
\begin{equation}
u-\mathbb{E}[u\mid x,x_{\mathcal B}]
\approx
J_z(x,x_{\mathcal B})\left(z-\mu_z(x,x_{\mathcal B})\right)
\in
\operatorname{Col}\!\left(J_z(x,x_{\mathcal B})\right),
\end{equation}
and
\begin{equation}
\operatorname{rank}\!\left(
J_z(x,x_{\mathcal B})\Sigma_z(x,x_{\mathcal B})J_z(x,x_{\mathcal B})^\top
\right)
\leq
\operatorname{rank}\!\left(J_z(x,x_{\mathcal B})\right)
\leq d_z.
\end{equation}
Although $\Sigma_z(x,x_{\mathcal B})$ is diagonal, the covariance entries
satisfy
\begin{equation}
\operatorname{Cov}(u_i,u_j\mid x,x_{\mathcal B})
\approx
\sum_{k=1}^{d_z}
J_{z,ik}(x,x_{\mathcal B})J_{z,jk}(x,x_{\mathcal B})\sigma_{z,k}^2(x,x_{\mathcal B}),
\end{equation}
which are generally nonzero for $i\neq j$. Although the decoder does not
explicitly depend on the molecular state, its Jacobian is evaluated at the
state-dependent latent mean $\mu_z(x,x_{\mathcal B})$. Consequently,
together with the state dependence of $\Sigma_z(x,x_{\mathcal B})$, the
induced covariance and its associated low-dimensional subspace may vary
with $(x,x_{\mathcal B})$.

\subsection{Algorithm and implementation details of FS-TPS and LaS-TPS}
The training and inference algorithms for FS-TPS and LaS-TPS are provided in
\textbf{Algorithm 1} and \textbf{Algorithm 2}. In the algorithms, \(x_{\mathcal B,t}^{(m)}
=\operatorname{Align}(x_t^{(m)},x_{\mathcal B})\) denotes the target
conformation \(x_{\mathcal B}\) re-expressed in the reference frame of the
current configuration \(x_t^{(m)}\), and
\(\xi_t^{(m)}=(x_t^{(m)},x_{\mathcal B,t}^{(m)})\) is the common policy
input used by both methods. The default sampled-policy evaluation uses \(\alpha=1\), corresponding to
the $\sigma$-scale factor reported in Table 1.
For FS-TPS, setting \(\alpha=0\) gives the conditional mean-force policy. For
LaS-TPS, it gives the decoded-latent-mean policy. \\

\paragraph{Detailed settings}
For the baselines, we use the default hyperparameter settings from TPS-DPS.
For alanine dipeptide, we use a batch size of 16, 1000 steps per path, an
annealing temperature from 600K to 300K, 1000 rollouts, and a buffer size of
1000. For chignolin, we use a batch size of 4, 5000 steps per path, an
annealing temperature from 600K to 300K, 100 rollouts, and a buffer size of
200. For BBL protein, we use a batch size of 2, 5000 steps per path, and an
annealing temperature from 600K to 400K.

For FS-TPS, $\beta_{\mathrm H}$ is set to $1\times10^{-5}$ for alanine
dipeptide, $5\times10^{-6}$ for chignolin, and $5\times10^{-5}$ for BBL
protein. For LaS-TPS, we evaluate
$\beta_{\mathrm{KL}}\in\{10^{-2},10^{-3},10^{-4}\}$ for alanine
dipeptide and chignolin, and
$\beta_{\mathrm{KL}}\in\{10^{-6},10^{-7}\}$ for BBL protein.
The architecture details are included in
Table~\ref{tab:network-architectures}. We use random seeds
$2, 50, 137, 4829, 44219, 72831, 560237, 910244, 8831021$ for alanine
dipeptide and $2, 42, 50, 137, 4829, 44219, 72831$ for chignolin and BBL
protein.

\paragraph{Replay buffer sampling}
For LaS-TPS, at each training step, we assemble a minibatch by drawing a small fraction
of trajectories from a low-RMSD pool and the rest uniformly from the full
replay buffer. Concretely, after each rollout, every trajectory's closest
approach to the target state is identified by a relaxed Gaussian indicator
with bandwidth $\sigma$, evaluated over the trajectory after Kabsch alignment.
The RMSD at that frame is stored alongside the trajectory in the buffer.
During training, $N_{\text{biased}}$ trajectories are sampled uniformly from
the bottom $p$-fraction of the buffer ranked by RMSD (the ``low pool''), and
the remaining trajectories are sampled
uniformly from the full buffer as usual. For example, for alanine dipeptide, we used $\sigma=0.1$, $p=0.3$, $N_{\text{batch}}=16$, and
$N_{\text{biased}}=1$, giving the reported
$\sim 1/16 \approx 6\%$ biased fraction.

\begin{table*}[t]
\centering
\caption{Network architectures used for TPS-DPS and LaS-TPS.}
\label{tab:network-architectures}
\small
\setlength{\tabcolsep}{6pt}
\renewcommand{\arraystretch}{1.0}

\begin{tabular}{lll}
\toprule
System & Method & Layer dimensions \\
\midrule

Alanine dipeptide
& TPS-DPS
& $(\mathrm{input},128,256,256,256,128,\mathrm{output})$ \\

& LaS-TPS
& \makecell[l]{
Encoder: $(\mathrm{input},64,128,64)$;\\
latent heads: $2\times(64,10)$;\\
decoder: $(10,64,128,64,\mathrm{output})$
} \\
\midrule

Chignolin
& TPS-DPS
& $(\mathrm{input},512,1024,2048,1024,512,\mathrm{output})$ \\

& LaS-TPS
& \makecell[l]{
Encoder: $(\mathrm{input},256,512,256)$;\\
latent heads: $2\times(256,24)$;\\
decoder: $(24,256,512,256,\mathrm{output})$
} \\
\midrule

BBL
& TPS-DPS
& $(\mathrm{input},512,1024,2048,1024,512,\mathrm{output})$ \\

& LaS-TPS
& \makecell[l]{
Encoder: $(\mathrm{input},1024,512,256)$;\\
latent heads: $2\times(256,128)$;\\
decoder: $(128,256,512,1024,\mathrm{output})$
} \\

\bottomrule
\end{tabular}
\end{table*}

\subsection{Evaluation Metrics}
Following evaluations in TPS-DPS~\cite{tpsdps}, for each sampled transition path, we roll out the trajectory under the learned
dynamics and record the atomic positions at every step, from which the
corresponding forces and potential energies are obtained by re-evaluating the
system's force field at each configuration. All metrics are always computed
over an ensemble of 64 sampled paths.

\paragraph{RMSD}Structural accuracy is quantified by the root-mean-square deviation (RMSD)
between the final configuration of each sampled path and the target basin's
reference structure, computed over heavy atoms only after optimal rigid-body
superposition via the Kabsch algorithm~\cite{kabsch1976}.

\paragraph{THP}Path success is defined through a target-hit criterion evaluated in a
low-dimensional collective-variable space specific to each system. For
alanine dipeptide, the backbone dihedral angles $(\phi,\psi)$ of the
final configuration are compared with those of the target state, and a path
is counted as successful if the periodic angular distance satisfies
$(\Delta\phi)^2+(\Delta\psi)^2<0.75^2$. For the larger proteins chignolin and
BBL, the final configuration is projected onto the leading two time-lagged independent
components (TIC1, TIC2) obtained from a TICA model~\cite{perezhernandez2013} trained on backbone-torsion
features of the folded reference structure. The same distance criterion,
$(\Delta\mathrm{TIC}_1)^2+(\Delta\mathrm{TIC}_2)^2<0.75^2$,
is applied in this reduced space. The target-hit percentage (THP) is the fraction of
sampled paths satisfying this criterion, expressed as a percentage.

\paragraph{ETS}For paths classified as successful, we additionally report the energy at the
transition state (ETS), defined as the maximum potential energy attained along
the trajectory. 

\paragraph{Mode coverage}For alanine dipeptide specifically, whose two-dimensional $(\phi,\psi)$ landscape admits
two known, well-separated first-order saddle points connecting the C5 and
C7ax basins, we further assess whether sampling explores both transition
mechanisms rather than collapsing onto a single dominant pathway. For each
successful path, we determine whether its trajectory passes within a fixed
radius ($0.3$ rad) of each saddle point in dihedral space at any point along
its length; paths that come within this radius of more than one saddle are
treated as ambiguous and excluded from this analysis.

\subsection{Computation of stochastic-force structure metrics}
\subsection{Per-state stochastic-force metrics}
Five per-state metrics including internal effective rank, internal top-mode explained
variance, excess cross-atom covariance, RMS stochastic amplitude relative to
the mean force, and latent effective-rank utilization, are computed from the
same per-state stochastic-force ensemble. The procedures are identical for
alanine dipeptide, chignolin, and BBL; only the number of atoms $N$, latent
dimension $d_z$, and number of evaluated states $S$ differ across systems.
We use $S=200$ states for alanine dipeptide and $S=224$ states for chignolin
and BBL.

Each metric is first computed independently at every fixed state. For each
trained seed, the resulting per-state values are then averaged over the state
bank. We finally report the mean and standard deviation across independently
trained seeds.

Throughout this section, the target conformation $x_{\mathcal B}$ is
held fixed across the entire state bank and is therefore omitted from
the notation; all quantities are written as functions of the
configuration $x$ alone.

\paragraph{Per-state force ensemble.}
For each fixed molecular configuration $x$, we
query the trained policy $K_{\mathrm{MC}}=1024$ times while holding $x$
unchanged, producing applied bias-force samples
\[
    \left\{\mathbf{f}^{(k)}(x)\right\}_{k=1}^{K_{\mathrm{MC}}},
    \qquad
    \mathbf{f}^{(k)}(x)\in\mathbb{R}^{3N}.
\]
For FS-TPS, each sample is drawn directly from the predicted force-space
Gaussian. For LaS-TPS, a fresh latent variable is sampled 
and subsequently decoded into the applied bias force. Let
\[
    \overline{\mathbf{f}}(x)
    =
    \frac{1}{K_{\mathrm{MC}}}
    \sum_{k=1}^{K_{\mathrm{MC}}}\mathbf{f}^{(k)}(x)
\]
denote the sample mean. The per-state force covariance is
\[
    \boldsymbol{\Sigma}(x)
    =
    \frac{1}{K_{\mathrm{MC}}-1}
    \sum_{k=1}^{K_{\mathrm{MC}}}
    \left(\mathbf{f}^{(k)}-\overline{\mathbf{f}}\right)
    \left(\mathbf{f}^{(k)}-\overline{\mathbf{f}}\right)^{\!\top}.
\]

\paragraph{Removal of rigid-body components.}
For the three internal-force metrics below, we remove the six-dimensional
rigid-body subspace consisting of three translational and three mass-weighted
rotational modes. Let $\mathbf{Q}(x)$ contain the corresponding basis vectors.
The projector onto the internal subspace is
\[
    \mathbf{P}_{\mathrm{int}}(x)
    =
    \mathbf{I}
    -
    \mathbf{Q}
    \left(\mathbf{Q}^{\top}\mathbf{Q}\right)^{-1}
    \mathbf{Q}^{\top},
\]
and the projected covariance is
\[
    \boldsymbol{\Sigma}_{\mathrm{int}}(x)
    =
    \mathbf{P}_{\mathrm{int}}
    \boldsymbol{\Sigma}(x)
    \mathbf{P}_{\mathrm{int}}^{\top}.
\]
Thus, the reported internal-force statistics measure only genuine molecular
deformations rather than global translation or rotation. For numerical stability, the covariance eigenvalues are computed as the
squared singular values of the projected, centered sample matrix divided by
$K_{\mathrm{MC}}-1$, rather than by explicitly forming and diagonalizing
the covariance matrix.

\paragraph{Internal effective rank.}
Let $\{\lambda_i(x)\}$ denote the nonzero eigenvalues of
$\boldsymbol{\Sigma}_{\mathrm{int}}(x)$, and define
\[
    p_i(x)=\frac{\lambda_i(x)}{\sum_j\lambda_j(x)}.
\]
The internal effective rank is the exponential of the spectral Shannon
entropy~\cite{roy2007effective},
\[
    r_{\mathrm{eff}}^{\mathrm{int}}(x)
    =
    \exp\left[-\sum_i p_i(x)\log p_i(x)\right].
\]
This quantity is a continuous measure of the effective number of
equally weighted stochastic-force directions. It equals $q$ when exactly
$q$ nonzero eigenvalues have equal magnitude.

\paragraph{Internal top-mode explained variance.}
The fraction of internal stochastic-force variance captured by the dominant
mode is
\[
    \mathrm{EV@1}(x)
    =
    \frac{\lambda_1(x)}{\sum_i\lambda_i(x)},
\]
where $\lambda_1$ is the largest eigenvalue. We report this value as a
percentage. Values near $100\%$ indicate that the stochastic force is
concentrated along a single internal direction, whereas smaller values
indicate a broader spectrum.

\paragraph{Excess cross-atom correlation.}
We partition $\boldsymbol{\Sigma}_{\mathrm{int}}$ into $N\times N$ blocks
$\boldsymbol{\Sigma}_{ab}\in\mathbb{R}^{3\times3}$, where each block describes
the covariance between the Cartesian force components of atoms $a$ and $b$.
The cross-atom covariance fraction is
\[
    R_{\mathrm{cross}}(x)
    =
    \frac{
        \sum_{a\neq b}\left\|\boldsymbol{\Sigma}_{ab}(x)\right\|_{F}^{2}
    }{
        \sum_{a,b}\left\|\boldsymbol{\Sigma}_{ab}(x)\right\|_{F}^{2}
    }.
\]
Because a finite number of Monte Carlo samples produces nonzero off-diagonal
covariance even for independent atom-wise noise, we estimate a permutation
null. Specifically, the $K_{\mathrm{MC}}$ sample indices are independently permuted for
each atom, which preserves each atom's marginal covariance while destroying
cross-atom coupling. Repeating this procedure $B=50$ times gives
\[
    R_{\mathrm{cross}}^{\mathrm{excess}}(x)
    =
    R_{\mathrm{cross}}^{\mathrm{real}}(x)
    -
    \frac{1}{B}\sum_{\ell=1}^{B}
    R_{\mathrm{cross}}^{\mathrm{null},\ell}(x).
\]
A value near zero therefore indicates no cross-atom structure beyond the
finite-sample background.

\paragraph{RMS stochastic amplitude relative to the mean force.}
This metric is computed from the full, unprojected covariance
$\boldsymbol{\Sigma}(x)$. The total stochastic variance is
$\operatorname{tr}\boldsymbol{\Sigma}(x)$, while the deterministic force
magnitude is $\|\overline{\mathbf{f}}(x)\|_2^2$. We define
\[
    R_{\mathrm{noise/mean}}(x)
    =
    \sqrt{
        \frac{
            \operatorname{tr}\boldsymbol{\Sigma}(x)
        }{
            \|\overline{\mathbf{f}}(x)\|_2^2
        }
    }.
\]
Values below one indicate that the stochastic component is smaller in RMS
magnitude than the mean bias force, whereas values above one indicate that
stochastic variation is the dominant contribution.

\paragraph{Latent effective-rank utilization.}
For LaS-TPS, we additionally verify that low-rank force covariance is not
caused by collapse of the sampled latent distribution. At each state, the
same $K_{\mathrm{MC}}=1024$ latent samples are used to construct the empirical latent
covariance $\boldsymbol{\Sigma}_{z}(x)$. Its effective rank is computed using
the same entropy-based definition above and normalized by the latent
dimension:
\[
    R_{z}(x)
    =
    \frac{r_{\mathrm{eff}}\!\left(\boldsymbol{\Sigma}_{z}(x)\right)}{d_z}.
\]
Values close to one indicate that the sampled latent noise remains
approximately full-rank. Consequently, a low-rank force covariance together
with $R_z\approx1$ reflects the structure induced by the nonlinear decoder,
rather than latent-space sampling collapse.

\subsection{Global effective rank across states}

Unlike the five metrics above, the global effective rank is computed jointly
across all states in the state bank rather than independently at each state.
For LaS-TPS, it measures whether the dominant internal stochastic-force
direction is shared across molecular states or varies along the transition
path. This metric is not reported for FS-TPS because its force covariance is
generally not characterized by a single dominant mode.

For each state \(x_s\) in the state bank, we obtain the unit-normalized leading
right singular vector
\[
    \mathbf{v}_1(x_s)\in\mathbb{R}^{3N}
\]
from the rigid-body-projected, mean-centered force sample matrix described
above. This vector represents the dominant internal stochastic-force direction
at that state.

Before comparing directions across states, the dominant force direction
at each state is expressed in the common target reference frame using the
Kabsch rotation that aligns the corresponding configuration $x_s$ to the
target conformation $x_{\mathcal B}$. 

We construct the cross-state directional matrix by assigning equal weight to
every state:
\[
    \mathbf{M}
    =
    \frac{1}{S}
    \sum_{s=1}^{S}
    \mathbf{v}_1(x_s)\mathbf{v}_1(x_s)^\top.
\]
Let \(\{\eta_i\}\) denote the eigenvalues of \(\mathbf{M}\), and define
\[
    q_i=\frac{\eta_i}{\sum_j\eta_j}.
\]
The global effective rank uses the same effective rank equation in former sections.
In practice, the spectrum is computed from the \(S\times S\) Gram matrix of
the aligned dominant modes, which has the same nonzero eigenvalues as
\(\mathbf{M}\) and avoids explicitly constructing a \(3N\times3N\) matrix.

A global effective rank near one indicates that the dominant stochastic-force
direction is approximately shared across all states. Larger values indicate
that the locally dominant direction changes across the state bank, even when
the force covariance at each individual state is approximately low-rank. The
metric is computed separately for each trained seed and then summarized by
the mean and standard deviation across seeds.

\subsection{Additional results and ablation studies}
\paragraph{Temperature-scaled Gaussian noise baseline}
For alanine dipeptide, we perturb the deterministic policy by directly adding
temperature-scaled Gaussian noise to the predicted bias force:
\[
\widetilde{\mathbf{u}}_{\theta}(x,T)
=
\mathbf{u}_{\theta}(x)
+
\sigma\sqrt{\frac{T}{T_{\mathrm{ref}}}}\,
\boldsymbol{\epsilon},
\qquad
\boldsymbol{\epsilon}\sim\mathcal{N}(\mathbf{0},\mathbf{I}),
\]
where $\sigma=10^{-3}$ and $T_{\mathrm{ref}}=300\,\mathrm{K}$.
Across nine seeds, this baseline achieves a mean THP of only $43.6\%$ and
remains strongly seed-sensitive, with THP ranging from $3.1\%$ to $70.3\%$,
indicating that simply adding Gaussian noise to a deterministic policy does not
reproduce the performance of FS-TPS.

\paragraph{Ablation of prioritized replay sampling} For alanine dipeptide, prioritized replay yields an average RMSD of
$0.25\pm0.03$~\AA{}, THP of $72.58\pm5.01\%$, and ETS of
$21.63\pm5.49$~kJ/mol across nine seeds, compared with
$0.25\pm0.04$~\AA{}, $71.30\pm8.24\%$, and
$18.54\pm7.06$~kJ/mol without prioritized sampling. (All results are reported across the same set of nine random seeds.) Thus, it provides only a
marginal increase in THP, while leaving RMSD essentially unchanged and not
improving ETS.

\paragraph{Effect of latent dimension.}
Table~\ref{tab:aldp-latent-dim} shows that reducing the latent dimension to one increases sensitivity to initialization for alanine dipeptide.

\begin{table}[!t]
\centering
\caption{
Effect of the latent dimension on LaS-TPS for alanine dipeptide.
Individual-seed results and mean $\pm$ standard deviation over four seeds are
reported. The seed exhibiting degraded performance is highlighted in red.
}
\label{tab:aldp-latent-dim}

\begingroup
\scriptsize
\setlength{\tabcolsep}{2.5pt}
\renewcommand{\arraystretch}{1.00}

\begin{adjustbox}{max width=\columnwidth,center}
\begin{tabular}{ccccc}
\toprule
\makecell{Latent\\dimension}
& Seed
& \makecell{RMSD\\(\AA{} $\downarrow$)}
& \makecell{THP\\(\% $\uparrow$)}
& \makecell{ETS\\(kJ/mol $\downarrow$)}
\\
\midrule

$d_z=1$  & 2    & 0.23 & 84.4 & 12.00 \\
         & 50   & 0.21 & 87.5 & 10.99 \\
         & 137  & 0.20 & 78.1 & 20.61 \\
\rowcolor{red!15}
         & \textbf{4829}
         & \textbf{0.70}
         & \textbf{21.9}
         & \textbf{60.29} \\
         & \textbf{Mean $\pm$ std.}
         & $0.34 \pm 0.24$
         & $67.98 \pm 30.96$
         & $25.97 \pm 23.28$
\\
\midrule

$d_z=10$ & 2    & 0.28 & 71.9 & 32.70 \\
         & 50   & 0.30 & 65.6 & 16.90 \\
         & 137  & 0.28 & 68.8 & 16.00 \\
         & 4829 & 0.25 & 70.3 & 19.10 \\
         & \textbf{Mean $\pm$ std.}
         & $0.28 \pm 0.02$
         & $69.15 \pm 2.68$
         & $21.18 \pm 7.79$
\\

\bottomrule
\end{tabular}
\end{adjustbox}
\endgroup
\end{table}

\subsection{Results of Doob's Lagrangian and FAS}
\begin{figure}[!t]
    \centering
    \hspace*{2mm}%
    \makebox[\columnwidth][c]{%
        \includegraphics[width=0.38\textwidth, trim=0 0.5cm 0 0.5cm, clip]
        {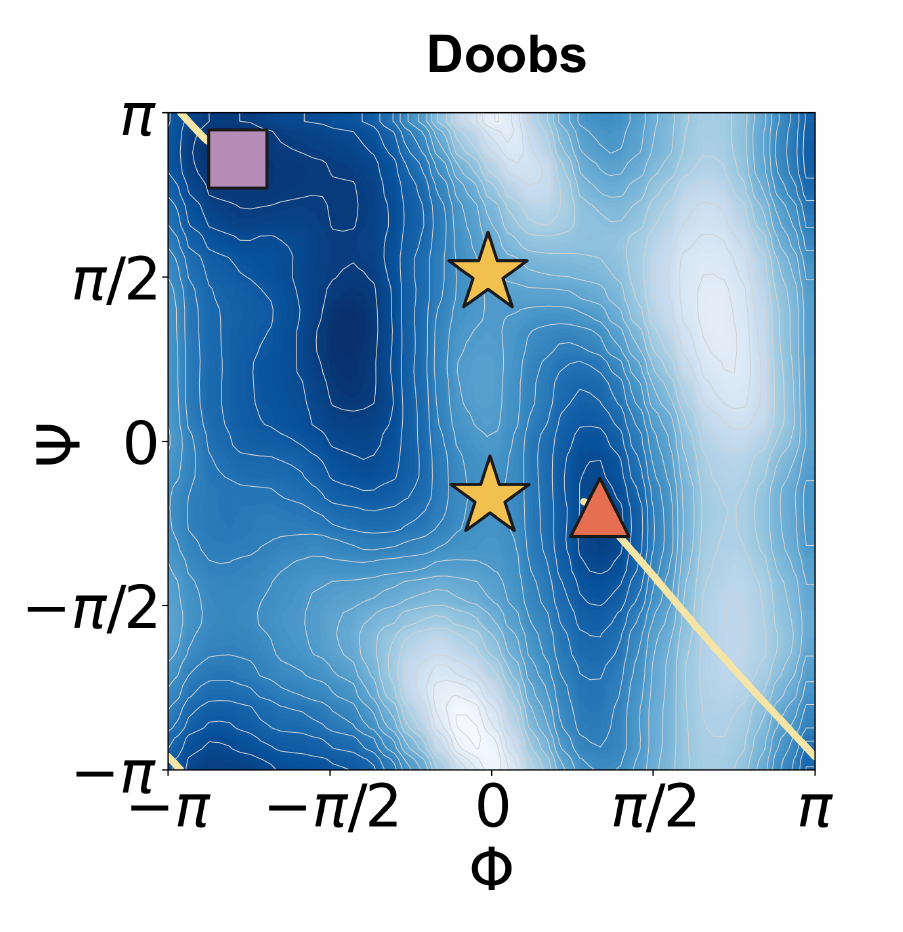}%
    }
    \caption{Sampled transition paths on the alanine
    dipeptide free-energy surface for Doob's Lagrangian.}
    \label{fig:doobs}
\end{figure}

We report the results of Doob's Lagrangian~\cite{doob} and FAS~\cite{fas} to conduct a thorough comparison between MD-rollout TPS methods and endpoint-conditioned path generators. The results are shown in Figure~\ref{fig:doobs} and Table~\ref{tab:endpoint-conditioned-comparison}. For  Doob's Lagrangian, all 64 paths collapse onto a single nearly straight trajectory in the $(\phi,\psi)$ collective-variable space.

\begin{table}[!t]
\centering
\caption{
Endpoint-conditioned path generators on alanine dipeptide
and chignolin.
These methods explicitly constrain the trajectory endpoints; therefore, their
RMSD and THP are determined primarily by the
endpoint conditions and are not directly comparable with those of unconstrained
transition path sampling methods.
Chignolin results of Doob's Lagrangian and all results for FAS are taken from Lee et al.~\cite{doobseq} and
Park et al.~\cite{fas}.
No corresponding results were reported for BBL.
}
\label{tab:endpoint-conditioned-comparison}

\begingroup
\scriptsize
\setlength{\tabcolsep}{2.5pt}
\renewcommand{\arraystretch}{1.05}

\begin{adjustbox}{max width=\columnwidth,center}
\begin{tabular}{@{}lccc@{}}
\toprule
Method
& \makecell{RMSD\\(\AA{} $\downarrow$)}
& \makecell{THP\\(\% $\uparrow$)}
& \makecell{ETS\\(kJ/mol $\downarrow$)}
\\
\midrule

\multicolumn{4}{c}{\textbf{Alanine Dipeptide (22 atoms)}}\\
\midrule

Doob's
& \multicolumn{2}{c}{\textbf{Endpoint-constrained}}
& $-17.20 \pm 0.01$
\\

FAS
& \multicolumn{2}{c}{\textbf{RMSD $\approx 0$, THP $\approx 100\%$}}
& $-29.46 \pm 3.87$
\\

\midrule
\multicolumn{4}{c}{\textbf{Chignolin (166 atoms)}}\\
\midrule

Doob's
& \multicolumn{2}{c}{\textbf{Endpoint-constrained}}
& $3828.38 \pm 0.10$
\\

FAS
& \multicolumn{2}{c}{\textbf{RMSD $\approx 0$, THP $\approx 100\%$}}
& $-360.85 \pm 53.71$
\\

\bottomrule
\end{tabular}
\end{adjustbox}
\endgroup
\end{table}

\clearpage
\begin{algorithm*}[t]
\caption{Training with FS-TPS (colored in yellow) and LaS-TPS
(colored in purple)}
\label{alg:stochastic-control-training}
\smallskip

\KwIn{
Number of rollouts $I$;
training steps per rollout $J$;
replay-buffer batch size $K$;
number of paths per rollout $M$;
target conformation $x_{\mathcal B}$;
annealing schedule
$\lambda_1=\lambda_{\mathrm{start}}
>
\cdots
>
\lambda_I=\lambda_{\mathrm{end}}$;
\colorbox{forceyellow}{%
    \strut force-entropy coefficient $\beta_{\mathrm H}$%
};
and
\colorbox{latentpurple}{%
    \strut latent KL coefficient $\beta_{\mathrm{KL}}$%
}
}

Initialize an empty replay buffer $\widehat{\mathcal D}$,
policy parameters $\theta$, and scalar parameter $w$\;

\smallskip

\For{$i=1,\ldots,I$}{

Generate $M$ paths
$\{x_{0:L}^{(m)}\}_{m=1}^{M}$
from biased MD simulations at annealing temperature $\lambda_i$.
At each MD step $t$, construct
\[
x_{\mathcal B,t}^{(m)}
=
\operatorname{Align}\!\left(
x_t^{(m)},
x_{\mathcal B}
\right),
\qquad
\xi_t^{(m)}
=
\left(
x_t^{(m)},
x_{\mathcal B,t}^{(m)}
\right),
\]
and apply one of the following stochastic control policies:\;

\smallskip

\par\noindent
\makebox[\linewidth][c]{%
\begin{minipage}[t]{0.47\linewidth}
\centering
\footnotesize

\colorbox{forceyellow}{%
    \strut\hspace{1.5mm}\textbf{FS-TPS}\hspace{1.5mm}%
}
\par\smallskip

\raggedright
Predict the conditional force distribution:
\[
\begin{aligned}
\pi_\theta\!\left(
\mathbf u_t
\mid
\xi_t^{(m)}
\right)
&=
\mathcal N\!\Bigl(
\boldsymbol{\mu}^u_\theta
    (\xi_t^{(m)}),
\\[-1mm]
&\qquad
\operatorname{diag}\!\bigl[
\boldsymbol{\sigma}^u_\theta
    (\xi_t^{(m)})^2
\bigr]
\Bigr).
\end{aligned}
\]

Draw fresh Gaussian noise:
\[
\boldsymbol{\epsilon}_t^{(m)}
\sim
\mathcal N(\mathbf 0,\mathbf I),
\]

and sample the bias force:
\[
\begin{aligned}
\mathbf u_t^{(m)}
&=
\boldsymbol{\mu}^u_\theta
    (\xi_t^{(m)})
\\[-1mm]
&\quad+
\boldsymbol{\sigma}^u_\theta
    (\xi_t^{(m)})
\odot
\boldsymbol{\epsilon}_t^{(m)}.
\end{aligned}
\]

\end{minipage}%
\hfill
\begin{minipage}[t]{0.47\linewidth}
\centering
\footnotesize

\colorbox{latentpurple}{%
    \strut\hspace{1.5mm}\textbf{LaS-TPS}\hspace{1.5mm}%
}
\par\smallskip

\raggedright
Predict the conditional latent distribution:
\[
\begin{aligned}
q_\theta\!\left(
\mathbf z_t
\mid
\xi_t^{(m)}
\right)
&=
\mathcal N\!\Bigl(
\boldsymbol{\mu}^z_\theta
    (\xi_t^{(m)}),
\\[-1mm]
&\qquad
\operatorname{diag}\!\bigl[
\boldsymbol{\sigma}^z_\theta
    (\xi_t^{(m)})^2
\bigr]
\Bigr).
\end{aligned}
\]

Draw fresh Gaussian noise:
\[
\boldsymbol{\epsilon}_t^{(m)}
\sim
\mathcal N(\mathbf 0,\mathbf I),
\]

and sample the latent control:
\[
\begin{aligned}
\mathbf z_t^{(m)}
&=
\boldsymbol{\mu}^z_\theta
    (\xi_t^{(m)})
\\[-1mm]
&\quad+
\boldsymbol{\sigma}^z_\theta
    (\xi_t^{(m)})
\odot
\boldsymbol{\epsilon}_t^{(m)}.
\end{aligned}
\]

Decode the sampled latent variable:
\[
\mathbf u_t^{(m)}
=
g_\theta\!\left(
\mathbf z_t^{(m)}
\right).
\]

\end{minipage}%
}

\par\smallskip

Update the replay buffer:
\[
\widehat{\mathcal D}
\leftarrow
\widehat{\mathcal D}
\cup
\{x_{0:L}^{(m)}\}_{m=1}^{M}.
\]

\For{$j=1,\ldots,J$}{

Sample a batch of $K$ paths
$\{x_{0:L}^{(k)}\}_{k=1}^{K}$
from $\widehat{\mathcal D}$\;

For each sampled path and step, construct
\[
x_{\mathcal B,t}^{(k)}
=
\operatorname{Align}\!\left(
x_t^{(k)},
x_{\mathcal B}
\right),
\qquad
\xi_t^{(k)}
=
\left(
x_t^{(k)},
x_{\mathcal B,t}^{(k)}
\right).
\]

Form the common log-variance objective:
\[
\mathcal L(\theta,w)
=
\frac{1}{K}
\sum_{k=1}^{K}
\left(
\log
\frac{
p_0\!\left(x_{0:L}^{(k)}\right)
\mathbf 1_{\mathcal B}\!\left(x_L^{(k)}\right)
}{
p_{\mathbf u_\theta}\!\left(x_{0:L}^{(k)}\right)
}
-
w
\right)^2.
\]

Augment the common objective with the regularizer corresponding
to the selected stochastic-control policy:\;

\noindent
\colorbox{forceyellow}{%
\parbox{\dimexpr\linewidth-2\fboxsep\relax}{%
\[
\mathcal L_{\mathrm{FS}}(\theta,w)
=
\mathcal L(\theta,w)
-
\frac{\beta_i}{K(L+1)}
\sum_{k=1}^{K}
\sum_{t=0}^{L}
\mathcal H\!\left[
\pi_\theta\!\left(
\cdot
\mid
\xi_t^{(k)}
\right)
\right],
\qquad
\beta_i
=
\beta_{\mathrm H}
\frac{\lambda_i}{\lambda_{\mathrm{end}}}.
\]
}}\;

\noindent
\colorbox{latentpurple}{%
\parbox{\dimexpr\linewidth-2\fboxsep\relax}{%
\[
\mathcal L_{\mathrm{LaS}}(\theta,w)
=
\mathcal L(\theta,w)
+
\frac{\beta_{\mathrm{KL}}}{K(L+1)}
\sum_{k=1}^{K}
\sum_{t=0}^{L}
D_{\mathrm{KL}}\!\left(
q_\theta\!\left(
\mathbf z_t
\mid
\xi_t^{(k)}
\right)
\,\middle\|\,
\mathcal N(\mathbf 0,\mathbf I)
\right).
\]
}}\;

Update $\theta$ and $w$ using either
$\nabla_{\theta,w}\mathcal L_{\mathrm{FS}}$
or
$\nabla_{\theta,w}\mathcal L_{\mathrm{LaS}}$\;

}
}

\end{algorithm*}

\label{app:inference-algorithm}

\begin{algorithm*}[t]
\caption{Inference with FS-TPS (colored in yellow) and LaS-TPS
(colored in purple)}
\label{alg:stochastic-control-inference}
\smallskip

\KwIn{
Trained policy parameters $\theta^\star$;
number of sampled paths $M_{\mathrm{eval}}$;
path length $L$;
reactant-state initialization distribution $\rho_{\mathcal A}$;
target conformation $x_{\mathcal B}$;
evaluation temperature $\lambda_{\mathrm{end}}$;
stochastic scale $\alpha$ (default $\alpha=1$)
}

\KwOut{
Sampled transition paths
$\mathcal{T}
=
\{\tau^{(m)}\}_{m=1}^{M_{\mathrm{eval}}}$
}

Initialize an empty path set
$\mathcal{T}\leftarrow\varnothing$\;

\smallskip

\For{$m=1,\ldots,M_{\mathrm{eval}}$}{

Initialize the molecular configuration and velocity:
\[
\left(
x_0^{(m)},v_0^{(m)}
\right)
\sim
\rho_{\mathcal A}.
\]

\For{$t=0,\ldots,L-1$}{

Construct the policy input:
\[
x_{\mathcal B,t}^{(m)}
=
\operatorname{Align}\!\left(
x_t^{(m)},
x_{\mathcal B}
\right),
\qquad
\xi_t^{(m)}
=
\left(
x_t^{(m)},
x_{\mathcal B,t}^{(m)}
\right).
\]
At the current configuration, apply one of the following
stochastic control policies:\;

\par\smallskip
\noindent
\hdashrule{\linewidth}{0.4pt}{3pt 2pt}
\par\smallskip

\makebox[\linewidth][c]{%
\begin{minipage}[t]{0.47\linewidth}
\centering
\footnotesize

\colorbox{forceyellow}{%
    \strut\hspace{1.5mm}\textbf{FS-TPS}\hspace{1.5mm}%
}
\par\smallskip

\raggedright
Predict the conditional force distribution:
\[
\begin{aligned}
\pi_{\theta^\star}\!\left(
\mathbf{u}_t
\mid
\xi_t^{(m)}
\right)
&=
\mathcal{N}\!\Bigl(
\boldsymbol{\mu}^{u}_{\theta^\star}
    (\xi_t^{(m)}),
\\[-1mm]
&\qquad
\operatorname{diag}\!\bigl[
\boldsymbol{\sigma}^{u}_{\theta^\star}
    (\xi_t^{(m)})^2
\bigr]
\Bigr).
\end{aligned}
\]

Draw fresh Gaussian noise:
\[
\boldsymbol{\epsilon}_t^{(m)}
\sim
\mathcal{N}(\mathbf{0},\mathbf{I}),
\]

and sample the bias force:
\[
\begin{aligned}
\mathbf{u}_t^{(m)}
&=
\boldsymbol{\mu}^{u}_{\theta^\star}
    (\xi_t^{(m)})
\\[-1mm]
&\quad+
\alpha\,
\boldsymbol{\sigma}^{u}_{\theta^\star}
    (\xi_t^{(m)})
\odot
\boldsymbol{\epsilon}_t^{(m)}.
\end{aligned}
\]

\end{minipage}%
\hfill
\begin{minipage}[t]{0.47\linewidth}
\centering
\footnotesize

\colorbox{latentpurple}{%
    \strut\hspace{1.5mm}\textbf{LaS-TPS}\hspace{1.5mm}%
}
\par\smallskip

\raggedright
Predict the conditional latent distribution:
\[
\begin{aligned}
q_{\theta^\star}\!\left(
\mathbf{z}_t
\mid
\xi_t^{(m)}
\right)
&=
\mathcal{N}\!\Bigl(
\boldsymbol{\mu}^{z}_{\theta^\star}
    (\xi_t^{(m)}),
\\[-1mm]
&\qquad
\operatorname{diag}\!\bigl[
\boldsymbol{\sigma}^{z}_{\theta^\star}
    (\xi_t^{(m)})^2
\bigr]
\Bigr).
\end{aligned}
\]

Draw fresh Gaussian noise:
\[
\boldsymbol{\epsilon}_t^{(m)}
\sim
\mathcal{N}(\mathbf{0},\mathbf{I}),
\]

and sample the latent control:
\[
\begin{aligned}
\mathbf{z}_t^{(m)}
&=
\boldsymbol{\mu}^{z}_{\theta^\star}
    (\xi_t^{(m)})
\\[-1mm]
&\quad+
\alpha\,
\boldsymbol{\sigma}^{z}_{\theta^\star}
    (\xi_t^{(m)})
\odot
\boldsymbol{\epsilon}_t^{(m)}.
\end{aligned}
\]

Decode the sampled latent variable into the bias force:
\[
\mathbf{u}_t^{(m)}
=
g_{\theta^\star}\!\left(
\mathbf{z}_t^{(m)}
\right).
\]

\end{minipage}%
}

\par\smallskip
\noindent
\hdashrule{\linewidth}{0.4pt}{3pt 2pt}
\par\smallskip

Combine the learned bias with the molecular force:
\[
\mathbf{F}_{t,\mathrm{total}}^{(m)}
=
\mathbf{F}_{\mathrm{MD}}\!\left(
x_t^{(m)}
\right)
+
\mathbf{u}_t^{(m)}.
\]

Advance the system by one biased-MD step:
\[
\left(
x_{t+1}^{(m)},
v_{t+1}^{(m)}
\right)
=
\operatorname{MDStep}_{\lambda_{\mathrm{end}}}
\!\left(
x_t^{(m)},
v_t^{(m)},
\mathbf{F}_{t,\mathrm{total}}^{(m)}
\right).
\]

}

Construct the sampled path:
\[
\tau^{(m)}
=
x_{0:L}^{(m)},
\]

and update the path set:
\[
\mathcal{T}
\leftarrow
\mathcal{T}
\cup
\{\tau^{(m)}\}.
\]

}

\Return{$\mathcal{T}$}\;

\end{algorithm*}


\end{document}